\documentclass[4pt]{article}

\usepackage{spconf,amsmath,graphicx}

\title{COUPLE LEARNING FOR SEMI-SUPERVISED SOUND EVENT DETECTION}

\twoauthors 
  {Rui Tao, Long Yan, Kazushige Ouchi}
	{ Toshiba China R\&D  Center, Beijing, China \\
	{ taorui, yanlong} @toshiba.com.cn, \\
	 kazushige.ouchi@toshiba.co.jp}
  {Xiangdong Wang}
	{ Beijing Key Laboratory of Mobile Computing \\
	 and Pervasive Device, Institute \\
	of Computing Technology,Chinese Academy \\
 	of Sciences, Beijing, China\\
    xdwang@ict.ac.cn}
    
\begin{document}
\maketitle

\begin{abstract}
The recently proposed Mean Teacher method, which exploits large-scale unlabeled data in a self-ensembling manner, has achieved state-of-the-art results in several semi-supervised learning benchmarks. Spurred by current achievements, this paper proposes an effective \textit{Couple Learning} method \footnote{code is available at https://github.com/Toshiba-China-RDC/dcase20\_task4/tree/master/PLG-MT\_run} that combines a well-trained model and a Mean Teacher model. The suggested pseudo-labels generated model (PLG) increases strongly- and weakly-labeled data to improve the Mean Teacher method's performance. Moreover, the Mean Teacher's consistency cost reduces the noise impact in the pseudo-labels introduced by detection errors. The experimental results on Task 4 of the DCASE2020 challenge demonstrate the superiority of the proposed method, achieving about 44.25\% F1-score on the validation set, significantly outperforming the baseline system’s 32.39\%. furthermore, this paper also propose a simple and effective experiment called the Variable Order Input (VOI) experiment, which proves the significance of the Couple Learning method. 
Our developed Couple Learning code is available on GitHub.
\end{abstract}
\section{Introduction}
\label{sec:intro}

Sound carries a large amount of information about our everyday environment and physical events. 
Therefore, developing signal processing methods to automatically extract this information has huge potential in several applications, such as searching for multimedia based on audio content by creating context-aware mobile devices, robots, and cars. 
Promoted by the annual DCASE challenges \cite{Mesaros2019,Turpault2020,Turpault2021}, the SOTA in weakly labeled semi-supervised sound event detection (SED) has progressed rapidly during recent years. Indeed, several approaches have been proposed for weakly labeled SED \cite{Jung2021,komatsu21_interspeech,deshmukh21_interspeech}, such as
An Integrated Pretrained Deep Neural Network \cite{Jung2021}, 
Classifier Chains \cite{komatsu21_interspeech}, 
and Self-Supervised Auxiliary \cite{deshmukh21_interspeech}. 
Most recent SOTA approaches, e.g., \cite{SzuYu2018,boes21_interspeech,Boes2021,Lin2019}, rely on neural attention, where the neural network performs audio tagging \cite{hong20_interspeech,chang20_interspeech} by learning to attend to the time range where the sound event is active. Then, the trained network is employed to locate sound events in time, although no strong labels have been used during training. Teacher-student approaches dominate Semi-supervised SED \cite{Chuming2020}, where the teacher and student networks are jointly trained to employ an additional loss for consistency between their predictions on unlabeled data.

Currently, several scholars separate the use of pseudo-label data from the Mean Teacher method, with some scholars focusing on improving the pseudo-labels data on semi-supervised tasks \cite{Sakiko2019,Kim2020,Ebbers2020}, while others focus on improving semi-supervised tasks utilizing the Mean Teacher method \cite{Turpault2019,zheng21_interspeech}. For example, \cite{Koh2021} employs the audio tagging system to generate pseudo labels with consistency training. Unlike current trends, this paper develops an ensemble method, entitled Couple Learning, which utilizes two deep learning models for semi-supervised sound event detection (SED). The two models have the distinctive objectives of generating pseudo-labels (PLG) from the unlabeled and weakly-labeled data in the original dataset and employing a Mean Teacher approach to predict sound classes using original data and the PLG’s outputs. A thorough description of the proposed method is given in Fig. 1.

Our architecture involves two main stages. The first stage employs a well-trained model to create pseudo weak and strong labels for the original unlabeled data and pseudo strong labels for the original weakly labeled data. The pseudo-labels generated at this stage are quite noisy. Therefore, the second stage utilizes the Mean Teacher model
to exploits the baseline and generates pseudo datasets.
The baseline dataset contains a few strongly labeled data and weakly labeled data, and a large number of unlabeled data. In contrast, the generated pseudo dataset contains pseudo-strong labels generated from unlabeled data (UPS), pseudo-weak labels generated from unlabeled data (UPW), and pseudo-strong labels generated from weakly-labeled data (WPS) data. 

\begin{figure*}
\centering
\includegraphics[width=0.9\textwidth]{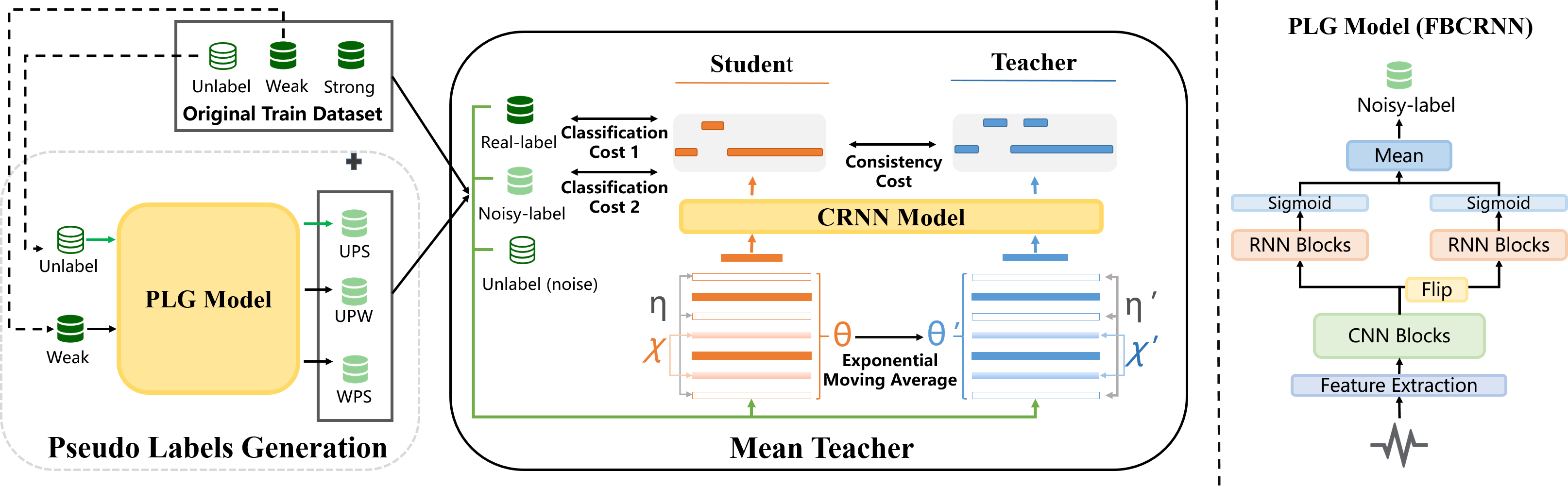}
\caption{\label{fig:frog} Illustration of Couple Learning method.}
\end{figure*}

\vspace{-0.5em}

\section{METHODOLOGY}
\label{sec:format}

\subsection{Couple Learning}
\label{ssec:subhead}

To overcome the Mean Teacher based CRNN model limitations, we propose the Couple Learning method to provide more useful information during model training. Specifically, the commonly used PLG model extracts information from raw data by generating pseudo-labels, but it introduces noise that negatively impacts the training process. Moreover, the Mean Teacher method involves only the noise suppression mechanism. However, a consistency cost can reduce the noise impact in the pseudo-labels due to detection errors by maintaining an exponential moving average of the label predictions on each training example and penalizing predictions inconsistent with this target. Therefore, the contribution of the proposed Couple Learning method is utilizing the information in the compound data more effectively. From this perspective, we couple the Mean Teacher method and the PLG model. 

Formally, we define the classification cost $J_{1}$ as the expected distance between the prediction (weight $\theta$, real-input $x$, and pseudo-input $\chi$) and the label (real-label $y$ and pseudo-label $y^{\prime}$). The classification cost of the Couple Learning method can be described as:

\begin{center}
\begin{equation}
J_{1}(\theta) =-\frac{1}{N} \{\sum\sum y log[f(x,\theta )]+ \sum\sum y^{\prime}  log[f(\chi ,\theta )]\} 
\end{equation}
\end{center}
Throughout the experiments, we employ cross-entropy as the classification cost.

The consistency cost $J_{2}$ is defined as the expected distance between the student model's prediction (weight $\theta$ and noise $\eta$) and the teacher model's prediction (weight $\theta^{\prime}$ and noise $\eta^{\prime}$). The consistency cost of the Couple Learning method is:

\begin{center}
\begin{equation}
J_{2}(\theta )=E_{x,\eta^{\prime} , \eta } \left [  \left \| f(x,\theta ^{\prime},\eta ^{\prime}  )- f(x,\theta ,\eta   )\right \|^{2}  \right ] 
\end{equation}
\end{center}
Next, the consistency cost function $J_{2}$ is approximated by sampling noise $\eta$, $\eta^{\prime}$ at each training step with stochastic gradient descent. The mean squared error (MSE) is the consistency cost utilized in the experiments \cite{14}.

\subsection{PLG model}
\label{ssec:subhead}
The PLG model is designed to help the Couple Learning method solve label scarcity by denoting the models that have been trained well and present a robust performance. Given that the greatest challenge in semi-supervised tasks is the limited number of labeled data that can be exploited to distill knowledge, the PLG model assists in extracting as much as possible knowledge from unlabeled or weakly labeled data. However, finding a suitable PLG model is critical. Generally, the PLG model can be a single network structure, an ensemble of multiple neural networks, or a rough manual labeling system. 
Nevertheless, this paper employs the Mean Teacher CRNN model (MT-CRNN), the baseline of the DCASE2020 challenge Task 4, utilzing a forward-backward convolutional recurrent neural network (FBCRNN) model and Hybrid ensemble model \cite{Ebbers2020} as the PLG models.
The labeled data generated by the PLG model contain much noise, even close to 50\%. 
The FBCRNN model, on the right side of Fig.1, was implemented based on Paderborn Sound Event Detection \cite{Ebbers2020}. It hypothesizes that the forward and backward layer jointly can get a better result at each time frame. This encourages the classifiers to capture sound events as soon as they appear.

\subsection{Mean Teacher CRNN model}
\label{ssec:subhead}
Inspired by the DCASE2020 Task 4 \cite{Turpault2019} baseline method, we utilize a Mean Teacher method and a CRNN model. During training, each batch contains a combination of unlabeled, weakly, and strongly labeled clips, with the system effectively using labeled, unlabeled, strongly-labeled, and weakly labeled data. The Mean Teacher method contains a student model and a teacher model, both employing a CNN block and two RNN blocks employing a bidirectional gated recurrent unit (GRU) scheme.

\subsubsection{Combination of labeled and unlabeled clips}
\label{sssec:subsubhead}
The system employs a Mean Teacher method that combines a student and a teacher model, sharing the same architecture. The student model is utilized at the final inference phase, and the teacher model guides the student model during training by calculating the exponential moving average of the student model’s weights that are exploited as the teacher's weights. This strategy allows exploiting both labeled and unlabeled data through consistency costs. Moreover, the student model is trained based on the labeled clips' classification costs, while as explained above, the teacher model is not trained, as its weights are the moving average of the student model's weight at each epoch. The teacher model receives the same input as the student model but with additive Gaussian noise from the unlabeled clips during training. This strategy helps train the student model via a consistency loss involving predictions of all clips in the batch.

\subsubsection{Combination of strongly-labeled and weakly-labeled clips}
\label{sssec:subsubhead}
The model combines a convolutional neural network (CNN) and a recurrent neural network (RNN), named CRNN, affording a unique cost calculation scheme that combines strongly- and weakly-labeled clips through two sets of parallel loss functions. The authors calculate the classification cost based on the binary cross-entropy between the student model and the labels, calculated at the frame level for the strongly-labeled clips and at the clip level for the weakly labeled clips. Simultaneously, the proposed method calculates the consistency cost based on the mean-squared error between the student and the teacher models. The mean-squared error is computed at the frame level for the strongly-labeled synthetic clips and the clip level for the weakly labeled clips.
\section{EXPERIMENTS}
\label{sec:pagestyle}
The following experiments aim to demonstrate the effectiveness of the proposed Couple Learning system on the DCASE 2020 Challenge Task 4 by involving three sets of experiments. Further details on the baseline system are presented in Section 2.3. 
\subsection{Dataset}
\label{ssec:subhead}
The experiments are conducted on the Task 4 benchmark datasets of the DCASE 2020 Challenge. The training datasets contain 14412 unlabeled in-domain training clips, 1578 weakly-labeled training clips with 2244 occurrences, and 2046 synthetic training clips with 6032 occurrences. The validation datasets involve 1168 clips with 4251 occurrences, while the public evaluation dataset contains 692 clips with 2765 occurrences - 10 classes of in-domain environment audio events, e.g., dog, dishes, and frying. The weakly labeled training data contains only utterance-level labels. Finally, the distribution per class of the in-domain unlabeled training set is closed to the labeled set.

\subsection{Evaluation metrics}
\label{ssec:subhead}

We evaluate our method's performance based on the event-based F1 score (EB-F1), the rank index of the DCASE2020 challenge Task 4. The EB-F1 metric compares the system's output and the corresponding reference on an event-by-event basis, measuring the system's ability to detect the correct event in the correct temporal position. Hence, the EB-F1 metric acts as an onset/offset detection capability measurement scheme.

\subsection{Setup}
\label{ssec:subhead}
To compare our proposed method against the baseline model, we employ mel-spectrograms with 128 mel bands for the input feature. The audio sampling rate is 16 kHz, and we extract the features utilizing the short-term Fourier transform coefficients (STFT) with 2048 sample windows and 255 sample hop sizes.

The CNN blocks involve seven stacked layers with [16, 32, 64, 128, 128, 128, 128] filters and  [[2, 2], [2, 2], [1, 2], [1, 2], [1, 2], [1, 2], [1, 2]] max-pooling per layer, with the convolution operations passing through a gated linear unit activation. The RNN block involves two stacked layers of 128 bidirectional GRU. The RNN block then passes through an attention pooling layer, which is the multiplication between a linear layer with a softmax activation and a linear layer with sigmoid activation.

\section{RESULT}
\label{sec:pagestyle}
\subsection{Contributions of the two stages }
\label{ssec:subhead}
The following experiments demonstrate the performance of the Mean Teacher method and PLG models on semi-supervised SED tasks. Specifically, Table 1 highlights that the EB-F1 of the CRNN without employing the Mean Teacher method is 28.14\%, while the EB-F1 increases to 32.39\% when the Mean Teacher method is employed. Moreover, the EB-F1 of the PLG model without using the Mean Teacher is 30.04\%, increasing to 33.93\% when the Mean Teacher method is used. The trend on the public evaluation set is the same as the validation set, proving that the Mean Teacher method and the PLG models contribute against higher performance.
More detailed experimental data are presented on GitHub.

\begin{table}[h]
	\begin{center}
		\begin{tabular}{|l|c|c|}
			\hline \bf Model & \bf EB-F1(\%)  \\ \hline
			CRNN & 28.14  \\
			+ Mean Teacher(Baseline) & 32.39  \\
			+ PLG  & 30.04  \\
			+ Mean Teacher + PLG & \textbf{33.93}\\
			\hline
		\end{tabular}
	\end{center}
	\caption{\label{font-table} EB-F1(\%) using different methods and introducing MT-CRNN as PLG on validation set. }
\end{table}

\vspace{-0.5em}

\begin{table}[h]
	\begin{center}
		\begin{tabular}{|l|c|c|}
			\hline \bf PLG Model  & \bf EB-F1(\%) \\ \hline
			Baseline  & 32.39 \\
			    + UPW  & 30.06 \\
			    + WPS  & 32.15 \\
			    + UPS  & 32.42 \\
			    + UPS + WPS  & 33.52 \\
			    + UPS + WPS + UPW  & \textbf{33.93} \\
			\hline
		\end{tabular}
	\end{center}
	\caption{\label{font-table} EB-F1(\%) introducing MT-CRNN as PLG with different pseudo-labels on validation set. }
\end{table}
\vspace{-0.5em}

\subsection{Contributions of the three kinds of pseudo-labels}
\label{ssec:subhead}
We further prove the performance of the PLG model on the semi-supervised SED task by performing several trials focusing on the interplay between the semi-supervised SED performance and employing pseudo-labels data. The corresponding results are presented in  Table 2. The original baseline method does not employ pseudo-label data, while adding pseudo-label data to the train set improves performance, as the information included in the data is effectively mined. Specifically, the performance improves when adding UPW, WPS, UPS individually, while the best performance is achieved when all these data types are used simultaneously. Based on these results, we conclude that pseudo labeled data is necessary for sound event detection. 

Table 2 reveals that employing the Mean Teacher CRNN (MT-CRNN) as the PLG model in the second column affords an EB-F1 value of 32.39\% for the baseline system, while this metric can reach 30.06\% when adding UPW data. Moreover, the EB-F1 metric increases to 32.15\% when WPS data are added and to 32.42\% when adding UPS data. All results are close to the baseline. The EB-F1 is up to 33.52\% when using UPS and WPS data to train the model, and adding all three data types during training increases the EB-F1 metric up to 33.93\%, i.e., 4.67\%  more than the baseline.

\subsection{Contribution of the Couple Learning method }
\label{ssec:subhead}
 Fig. 2 illustrates that when the Couple Learning system employs the Hybrid ensemble model \cite{Ebbers2020} as PLG, it achieves the best performance. Furthermore, using the FBCRNN model as PLG is the second-best, and employing the MT-CRNN as PLG is the third-best performing architecture. It should be noted that all three methods used as PLG achieve performances higher than the baseline. 
 
The Variable Order Input (VOI) experiment verify the impact of employing various input-orders of real-labels and pseudo-labels in an epoch. Real first (RF) means that all real-label data are arranged before the pseudo-labels data in an epoch. Pseudo first (PF) means that all pseudo-label data are arranged before the real-labels data in an epoch, while random means that all pseudo-labels and real-labels data are randomly arranged in an epoch. In other words, PF and RF employ the Couple Learning method only in a part of each epoch, and the random case employs the Couple Learning method during the entire epoch.

The random case fully mixes the unlabeled and pseudo-labeled data during each batch, presenting the best performance on the three PLG models. Additionally, PF is better than RF when Hybrid is employed as a PLG model, RF is better than PF when FBCRNN is used, and RF is better than PF when the MT-CRNN is employed as the PLG model. These experimental results further prove that the consistency cost has a noise suppression capability on the pseudo-labeled data employed in the Couple Learning method.

\begin{figure}
\centering
\includegraphics[width=0.48\textwidth]{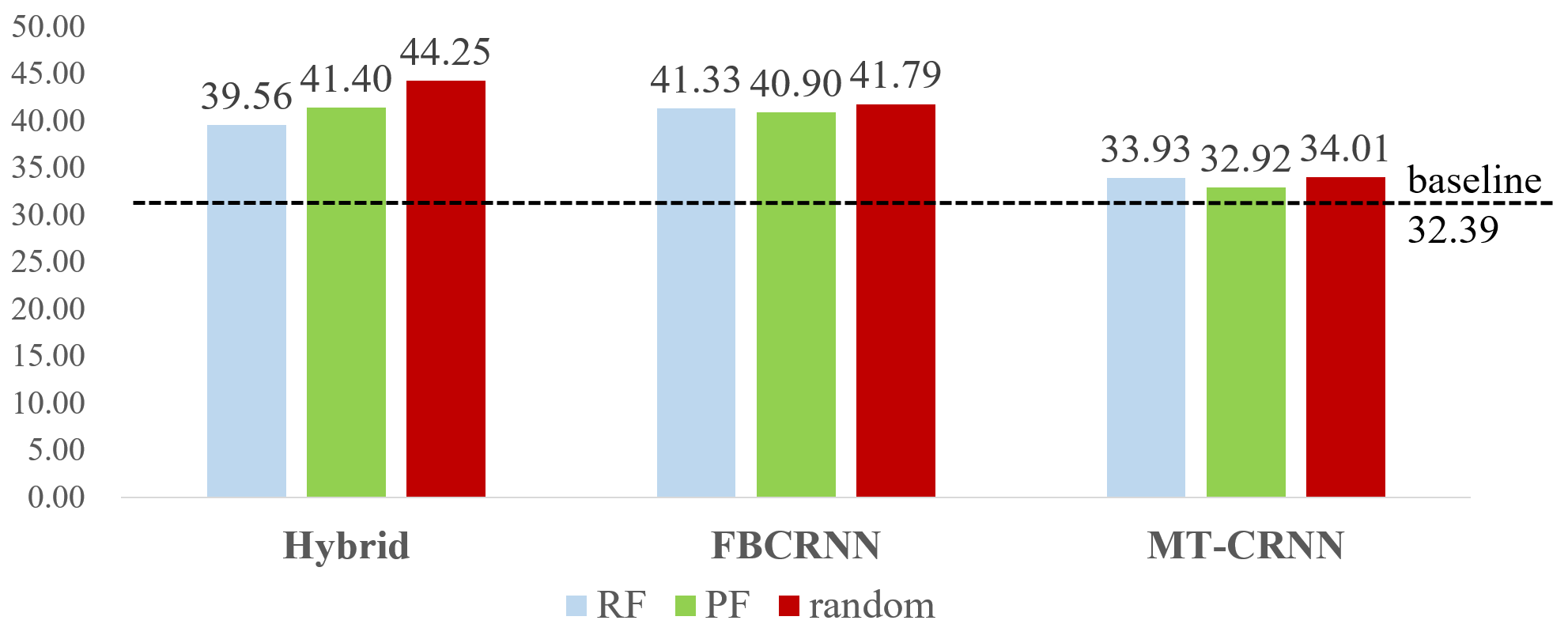}
\caption{\label{fig:frog}  EB-F1(\%) with different PLG models and three types of input-order (RF, PF, random) on validation set.}
\end{figure}
\vspace{-0.5em}

\section{CONCLUSION}
\label{sec:majhead}

This paper proposes the Couple Learning method, which is appropriate for semi-supervised tasks. Our method is based on the DCASE2020 Task 4 baseline, which involves a Mean Teacher method with a CRNN model, and we incorporate a PLG model to improve performance further. Our experiments demonstrate that the PLG model and the Mean Teacher method improve the system's performance, and their combination affords an even greater improvement. Therefore, this work concludes that the Mean Teacher and pseudo-labels collaborate well, improving performance. Furthermore, the Couple Learning method affords utilizing two model structures. Moreover, this paper proves that the Couple Learning method is a simple and powerful solution by VOI experiment.
Future work will aim to verify the Couple Learning method's effectiveness on more semi-supervised tasks,such as object detection tasks in the vision domain.

\bibliographystyle{IEEEtran}
\bibliography{mylib}

\end{document}